# Reblur2Deblur: Deblurring Videos via Self-Supervised Learning

Huaijin Chen[1,2]    Jinwei Gu[1]    Orazio Gallo[1]    Ming-Yu Liu[1]    Ashok Veeraraghavan[2]    Jan Kautz[1]

[1]{jinweig, ogallo, mingyul, jkautz}@nvidia.com   [2]{huaijin.chen, vashok}@rice.edu

[1] NVIDIA Research    [2] Rice University

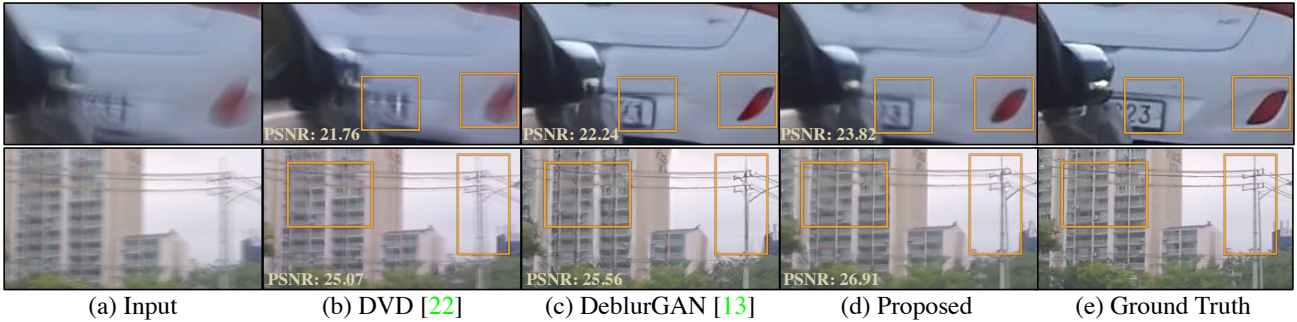

| (a) Input | (b) DVD [22] | (c) DeblurGAN [13] | (d) Proposed | (e) Ground Truth |

Figure 1. We propose a novel method for video motion deblur with self-supervised learning. Compared to prior method such as DVD [22] and DeblurGAN [13], we enforce a physics-based blur formation model during Deep Neural Network (DNN) learning, which effectively reduces image artifacts and improves generalization ability of DNN-based video deblurring.

## Abstract

*Motion blur is a fundamental problem in computer vision as it impacts image quality and hinders inference. Traditional deblurring algorithms leverage the physics of the image formation model and use hand-crafted priors: they usually produce results that better reflect the underlying scene, but present artifacts. Recent learning-based methods implicitly extract the distribution of natural images directly from the data and use it to synthesize plausible images. Their results are impressive, but they are not always faithful to the content of the latent image. We present an approach that bridges the two.*

*Our method fine-tunes existing deblurring neural networks in a self-supervised fashion by enforcing that the output, when blurred based on the optical flow between subsequent frames, matches the input blurry image. We show that our method significantly improves the performance of existing methods on several datasets both visually and in terms of image quality metrics.*

## 1. Introduction

Cameras integrate the scene radiance over a *finite* exposure time, which causes motion blur when the scene, the camera itself, or both move. Motion blur, in turn, affects the frequency content of the resulting image, thus hindering virtually any computer vision task. As a consequence, a number of deblurring algorithms have been proposed, which seek to estimate the latent, sharp image from one or more blurry observations of a scene.

At high level, we can identify two classes of deblurring algorithms. *Physics-based methods* observe that blur can be modeled by a convolution of the latent image with a spatially varying filter, usually referred to as the blur kernel. Deblurring, then, is reduced to estimating the blur kernel and deconvolving it from the observed, blurry image. Both the kernel estimation and the deconvolution process, however, are severely ill-posed and introduce artifacts, such as ringing artifacts [21].

A more recent trend is to use *deep-learning methods* to synthesize, from blurry observations, an image that best reflects the priors learned from training data [22, 13]. While neural networks have shown superior results in several fields of computer vision, they may require more training data than is available, and do become brittle when the training examples fail at capturing the full distribution of the real-world data. Finally, and perhaps most importantly, the synthesized images may be aesthetically pleasing, *e.g.*, sharper than the observations, but may not match the appearance of the latent image. Figure 1(c) shows one such example: the method by Kupyn *et al.* [13], reconstruct a sharp number "1," in place of the true number "3." This significantly reduces the benefit of using deep-learning based deblurring as a pre-processing step to computer vision algorithms.

Our method bridges the two approaches. Like previous methods, we use deep learning to synthesize sharp frames from a blurry video. However, we also explicitly enforce the solution to lie on the manifold of the sharp images that explain the observed blurry frames. To achieve this, we propose the first self-supervised, end-to-end deblurring method. Our differentiable pipeline can be used to fine-tune any existing pre-trained network.

From multiple consecutive blurry video frames, we produce the corresponding sharp frames and the optical flow between them. We use this information to compute a per-pixel blur kernel with which we *reblur* the sharp frames back onto the input images. By minimizing the distance between the synthesized blurry images and the input images, our approach allows to fine-tune the parameters of our system for specific inputs and *without the need for the corresponding ground truth data*. We show that our method outperforms the state-of-the-art methods based on both image quality metrics and visual inspection. Figure 1 shows two examples on which existing methods either fail or synthesize sharp estimates that do not reproduce the actual content of the latent sharp image. On the contrary, our method successfully recovers the ground truth image, while deblurring the input image.

## 2. Related Work

Traditionally, deblurring algorithms model the image formation process as the convolution of a blur kernel with a sharp image, which is then estimated by means of deconvolution [15].

The blur kernel, however, is anything but straightforward to estimate. A common assumption is that the kernel is space-invariant [6, 21, 15], which is only valid when the scene is static—camera shake is the only source of motion blur—and planar. Even under this simplifying assumption, the problem is severely ill-posed, and thus requires regularization. Indeed several priors have been successfully employed for the latent image, *e.g.*, TV [1], heavy-tailed gradient distribution [19], Gaussian distribution [14], smoothness [21], and for the shape of the kernel, *e.g.*, sparsity of the kernel [6] or parametric kernel modeling [27]. Alternatively, the kernel can be estimated with a deep-learning approach [20, 2]. A few methods relax this assumption by estimating non-uniform blur kernels [25, 9, 24]. Several approaches move even further in this direction by leveraging optical flow to estimate a *per-pixel* blur [10, 7].

The strength of these methods lies in their ability to explicitly model the physics of the image formation model. On the other hand, hand-crafted priors are not always realistic, and may result in visual artifacts in the estimated image.

An alternative way to tackle this problem is to learn directly from the data, which is possible thanks to the success of learning-based image synthesis approaches. Rather than estimating the blur kernel and explicitly deconvolving it from the observed image, these methods propose an end-to-end learning strategy to estimate the latent image.

To compensate for the lack of priors, some methods rely on video data [22, 26, 3]: because camera shake is an idiosyncratic motion, each frame can be thought as an independent observation of the scene.

In the absence of temporal information, a way to learn the priors is needed: Generative Adversarial Networks (GANs) are a powerful method for this task. Indeed a number of approaches successfully employ GANs to perform single-image deblurring [13, 18, 17].

Standard GAN-based methods have shown an extraordinary ability to learn natural distributions. However, the images they synthesize, while realistic, may not reflect the input data accurately, as shown in Figure 1(c).

We propose to leverage the benefit of both approaches: after a kernel-free estimation of the latent image from multiple frames, we estimate the optical flow, and the per-pixel blur it induces. This allows us to *reblur* the estimated sharp images back on the observed blurry images, and to constrain the solution to the manifold of images that capture the *actual* input content.

## 3. Method

Our work stems from the observation that GAN-based methods produce excellent results thanks to their ability to capture the natural distribution of sharp images. However, we also note that the synthesized images, while generally sharp, may deviate from the content of the latent image that underlying the blurry observation, see Figure 1(c).

We propose to improve the performance of existing end-to-end deblurring methods by encouraging solutions that more closely capture the content of the input image, in addition to being sharp. Our key idea is to introduce a physics-based blur formation model into the DNN training, with which we *reblur* the estimated sharp images and minimize the difference with the input blurry images.

### 3.1. Overview and Notation

Given a deblur network $d(\cdot; \Theta_d)$, either pre-trained or trained from scratch, we estimate the latent sharp frame at time $t$, from the blurry observation $I_B^{(t)}$:

$$\hat{I}_S^{(t)} = d(I_B^{(t)}; \Theta_d). \qquad (1)$$

The weights of the deblurring network $\Theta_d$ can be learned by minimizing the loss $\mathcal{L}_\mathcal{S}$ over a dataset $\mathcal{S} = \{I_B; I_S\}$ with supervision

$$\mathcal{L}_\mathcal{S}(\Theta_d) = \sum_\mathcal{S} h(\hat{I}_S, I_S), \qquad (2)$$

where $h(\cdot, \cdot)$ measures the distance between the estimated sharp images and the ground truth sharp images. The supervised loss $\mathcal{L}_\mathcal{S}(\Theta_d)$ can be computed with different choices of the distance function $h(\cdot, \cdot)$ [28], or with multiple input frames [22], or at multiple scales [16]. Recent work [18, 13, 16] introduced an additional GAN loss $\mathcal{L}_\mathcal{G}(\Theta_d)$, which achieved better performance by implicitly learning the distribution of sharp images. Relying solely on supervised learning, these methods still often produce image artifacts, especially when they are applied to images with different distributions of the training images.

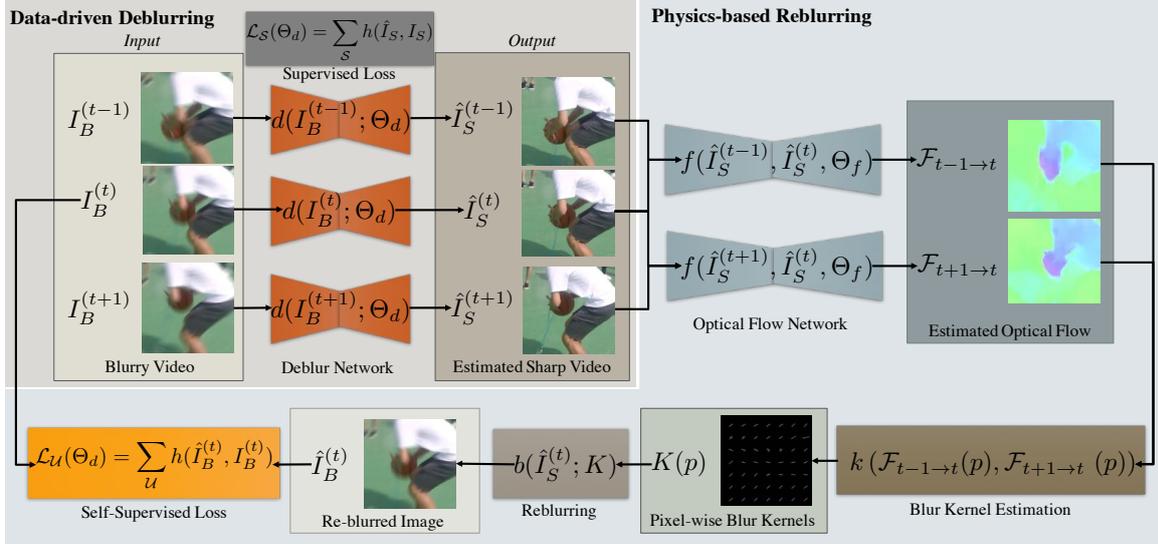

Figure 2. **System Overview**. Our proposed deblurring framework takes three consecutive blurry images as inputs. We first deblur each input image through the deblur network. After that, we compute the optical flow between the three recovered sharp images. We then estimate the per-pixel blur kernel and reconstruct the blurry input — this "reblurring" step offers an additional training signal for self-supervised learning to improve the deblur network and remove image artifacts.

Assume now that we are given a motion blurred video, rather than a single frame. By exploiting the motion information from videos, we incorporate a physics-based blur formation model into DNN learning to minimize image artifacts. Specifically, suppose we deblur three consecutive frames independently, obtaining $\{\hat{I}_S^{(t-1)}, \hat{I}_S^{(t)}, \hat{I}_S^{(t+1)}\}$. Since the frames are adjacent in time, we can compute the optical flow $\mathcal{F}$ between them. For this task we can use a different pre-trained neural network $f$:

$$\mathcal{F}_{t-1 \to t} = f(\hat{I}_S^{(t-1)}, \hat{I}_S^{(t)}, \Theta_f) \quad (3)$$
$$\mathcal{F}_{t+1 \to t} = f(\hat{I}_S^{(t+1)}, \hat{I}_S^{(t)}, \Theta_f). \quad (4)$$

Recall that our method aims at removing *motion* blur: the optical flow $\mathcal{F} = (u, v)$, which expresses the motion at each pixel $p$, carries the necessary information to estimate a per-pixel blur kernel

$$K(p) = k\left(\mathcal{F}_{t-1 \to t}(p), \ \mathcal{F}_{t+1 \to t}(p)\right), \quad (5)$$

which serves an important function: it allows to close the loop with the original observation. If image $\hat{I}_S^{(t)}$ is estimated correctly, in fact, the corresponding reblurred image,

$$\hat{I}_B^{(t)} = b(\hat{I}_S^{(t)}; K), \quad (6)$$

should be close to the input blurry image $I_B^{(t)}$, where $b(\cdot; K)$ is the physics-based blur formation model.

This pipeline allows us to fine-tune the deblur network $\Theta_d$ over new blurry videos $\mathcal{U} = \{I_B^{(t)}\}$ by minimizing [1]

$$\mathcal{L}_\mathcal{U}(\Theta_d) = \sum_\mathcal{U} h(\hat{I}_B^{(t)}, I_B^{(t)}), \quad (7)$$

where $\mathcal{L}_\mathcal{U}(\Theta_d)$ is the self-supervised loss over the unlabeled videos $\mathcal{U}$. Equation (7) is precisely the mechanism by which we enforce that the estimated image $\hat{I}_S^{(t)}$ is consistent with the observed image $I_B^{(t)}$, in addition to following the distribution of natural, sharp images.

We implement all the three modules, *i.e.*, the debur network $d(\cdot; \Theta_d)$, the optical flow network $f(\cdot; \Theta_f)$, and the physics-based blur formation model $b(\cdot; K)$, to be differentiable, which allows an end-to-end training. Figure 2 shows the complete system pipeline. In the following sections we describe the implementation details of each module.

## 3.2. Deblur Network and Optical Flow Network

Our system can flexibly choose the deblur network $d(\cdot; \Theta_d)$ or the optical flow network $f(\cdot; \Theta_f)$. In this paper, for the deblur network, we evaluated two network architecture. The first one is DVD [22], which is an encoder-decoder architecture with skip connections. The second one is the generator part of DeblurGAN [13] which is also an encoder-decoder architecture, but with a global skip connection between the input layer and output layer to learn the image residuals, and 9 "ResBlocks" [8] in the middle latent representation layers. For the optical flow network, we used FlowNetS [5] given its simplicity. Other variants can also be used, such as [23] and [11]. For the function $h(\cdot, \cdot)$ that measures the distance between two images, we used the MSE distance in the paper.

---
[1] We can also simultaneously fine-tune the optical flow network $\Theta_f$ together with $\Theta_d$.

### 3.3. From Optical Flow to Reblurring

The physics-based blur formation model $b(\hat{I}_S; K)$ performs a per-pixel convolution with a spatially-varying blur kernel $K(p)$ that is derived from the computed the optical flow $\mathcal{F}_{t-1 \to t}(p) = [u_{t-1 \to t}(p), v_{t-1 \to t}(p)]$ and $\mathcal{F}_{t+1 \to t}(p) = [u_{t+1 \to t}(p), v_{t-1 \to t}(p)]$, We assumed the same piecewise-linear motion blur kernel as proposed in [10] that consists of two line segments:

$$K(p)[x,y] = \begin{cases} \frac{\delta(-xv_{t+1\to t}(p)+yu_{t+1\to t}(p))}{2\tau||\mathcal{F}_{t+1\to t}(p)||} & \text{if } (x,y) \in R_1 \\ \frac{\delta(-xv_{t-1\to t}(p)+yu_{t-1\to t}(p))}{2\tau||\mathcal{F}_{t-1\to t}(p)||} & \text{if } (x,y) \in R_2 \\ 0 & \text{otherwise,} \end{cases} \quad (8)$$

where $R_1 : x \in [0, \tau u_{t+1 \to t}(p)], y \in [0, \tau v_{t+1 \to t}(p)]$ and $R_2 : x \in [0, \tau u_{t-1 \to t}(p)], y \in [0, \tau v_{t-1 \to t}(p)]$. $\tau$ is the exposure cycle as defined in [10], which is set to $\tau = 1$ in this paper.

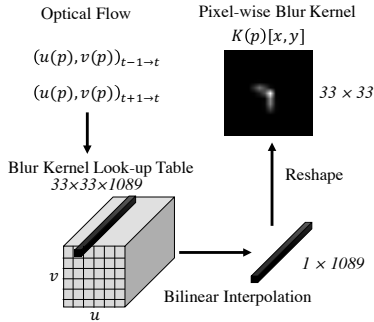

Figure 3. **Optical Flow to Blur Kernel:** We use a pre-computed lookup table and bilinear interpolation to convert the optical flow $(u(p), v(p))$ to per-pixel blur kernel $K(p)[x, y]$. This operation is differentiable, therefore can be instantiated as DNN layers.

The per-pixel blur kernel model defined above in Equation (8) cannot be directly used in DNN training, because the delta function $\delta(\cdot)$ is non-differentiable. To solve this issue, we use a precomputed lookup table that maps a set of optical flow values $(u_i, v_i)$ to the blur kernels $k_i[x, y]$, where $i = 1, \cdots, N$. For a given optical flow $(u, v)$ at pixel $p$, we use bilinear interpolation to compute the blur kernel $K(p)[x, y]$ from the lookup table:

$$K(p)[x,y] = \sum_{i=1}^{N} \omega_i(u,v) k_i[x,y]. \quad (9)$$

Since the bilinear interpolation is differentiable with respect to the weights $\omega_i(u, v)$, the gradient can be back-propagated to the optical flow network $f(\cdot; \Theta_f)$ and, subsequently, to the deblur network $d(\cdot; \Theta_d)$. In this paper, we set $N = 33 \times 33$ to compute the lookup table, thus limiting the range of the optical flow to compute the motion blur kernels from $-16$ pixels to 16 pixels in both directions. Figure 3 shows a diagram of this procedure.

### 3.4. Other Implementation Details

For the fine-tuning step, we found that minimizing the self-supervised loss $\mathcal{L}_\mathcal{U}$ alone leads to degenerate solutions. Therefore, we use the hybrid loss

$$\mathcal{L}(\Theta_d) = \mathcal{L}_\mathcal{S}(\Theta_d) + \alpha \mathcal{L}_\mathcal{U}(\Theta_d), \quad (10)$$

which also includes the supervision loss $\mathcal{L}_\mathcal{S}$ from the original supervised datasets $\mathcal{S}$. Each mini-batch is sampled partially from the original supervised datasets $\mathcal{S}$ and partially from the new unlabeled videos $\mathcal{U}$. The weight coefficient $\alpha$ balances the contribution of the self-supervision loss and the supervised loss. We set $\alpha = 0.1$ in all our experiments.

We implemented our algorithm in PyTorch. For both the deblur network $d(\cdot; \Theta_d)$ and the optical flow network $f(\cdot; \Theta_f)$, we started from pre-trained models, which we refer to as baseline networks. We performed 200 self-supervised learning iterations on the baseline networks. We used the ADAM solver [12] with a learning rate of 0.0001 for all our experiments, and a learning rate decay of 0.5 applied at 30, 50, and 100 epochs. We set the training mini-batch size between 8 and 20 depending on the DNN memory footprint. Finally, we use NVIDIA TitanX GPUs for training and testing.

Figure 4 shows an example of the proposed self-supervised learning. By fine-tuning with the physics-based blur formation model, we improve both the deblur network $\Theta_d$ and the optical flow network $\Theta_f$. The blur in this example is caused only by camera motion, and thus the true optical flow should be smooth — the fine-tuning removes the artifacts due to the scene's texture from the original optical flow. The proposed DeblurGAN-Reblur also outperforms the baseline DeblurGAN.

## 4. Experiments and Results

In this section we describe the experiments we performed, including the datasets we used, our quantitative and qualitative results, as well as a thorough ablation study. More results, including full size images are available in the supplementary material.[2]

---
[2]Supplementary material: https://goo.gl/nYPjEQ

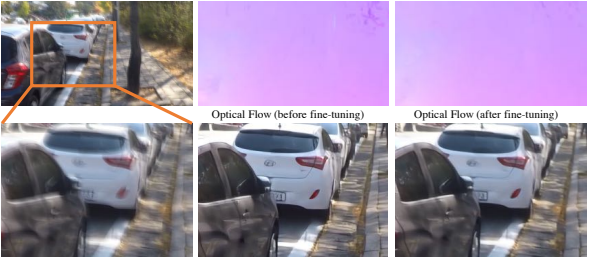

Figure 4. **A closer look of the proposed self-supervised learning results**. By fine-tuning via the physics-based blur formation model, we improve both the deblur network $\Theta_d$ and the optical flow network $\Theta_f$. The blur in this example image is caused only by camera motion — the fine-tuning removes the artifacts in the optical flow (due to scene texture). The deblurred image of DeblurGAN-Reblur is also better than the baseline DeblurGAN.

Table 1. Datasets used in our self-supervised fine-tuning experiments

| Name | # Images Used | Description |
| --- | --- | --- |
| MCD [16] | 1111 | GoPro, natural scenes |
| DVD [22] | 150 | GoPro/etc. cameras, natural scenes |
| ISOCHART | 129 | GoPro, ISO-Resolution chart |
| WFA [4, 3] | 150 | Blurry images only, natural scenes |

### 4.1. Datasets and Baselines

We evaluated the proposed method extensively on four datasets, as summarized in Table 1. Both the MCD [16] and DVD [22] datasets were captured with high-speed camera such as GoPro Hero 5 and Sony RX10 at 240fps, and thus have ground truth sharp images for quantitative evaluation. In addition, in order to test the generalization ability of deblurring algorithms (other than natural scenes), we also used a GoPro camera and captured a small dataset of an ISO-resolution chart moving in front of the camera. We refer to this as the ISOCHART set, which we will release upon publication. For these three datasets, we average every 9 frames to create the blurry image and use the center frame as the sharp ground truth. Finally, the WFA dataset [4, 3] is widely used for evaluating deblurring algorithms. It does not offer ground truth images and thus can only be evaluated qualitatively. Note that all these four datasets are used as the unsupervised dataset $\mathcal{U}$ in our experiment, which means we use only the blurry images as the input for self-supervised learning.

As mentioned early, we compared with two recent methods as our baseline, the DVD [22], which is an auto-encoder with skip connection for deblurring, and the DeblurGAN [13], which is the generator part of a GAN network. These two methods are representative since one is purely supervised learning from blur-sharp pairs and the other incorporates the GAN loss. We applied the proposed self-supervised learning method on top of these two baselines and fine-tuned the networks. We refer to the resulting networks as DVD-Reblur and DeblurGAN-Reblur respectively. In addition, we also compared with the MCD [16] method, which is similar to DeblurGAN.

### 4.2. Results

Figure 5 shows several examples from the four datasets. Table 2 shows the averaged PSNR and SSIM for the three datasets. As shown, our proposed self-supervised learning method brings significant improvement over the two baseline networks (especially over the DeblurGAN-baseline, about 1 dB improvement). Both proposed methods DeblurGAN-Reblur and DVD-Reblur effectively remove the artifacts introduced by the networks by enforcing the physics-based blur formation model.

We also compared with the MCD network [16], which is a multi-scale network with a GAN loss. The average PSNR of MCD on the three datasets are 28.53, 31.21, and 32.30 respectively, which are slightly better than ours (DeblurGAN-Reblur). However as shown in Figure 5 and the supplementary material, we found DeblurGAN-Reblur often achieves better visual quality than MCD (see the ISOCHART in Figure 5 for example). In addition, our proposed methods are computationally more efficient than MCD: MCD takes 4.33 seconds to deblur an image at resolution $1280 \times 720$, while DeblurGAN-Reblur takes 0.85 second and DVD-Reblur taks 0.84 second to deblur at the same image resolution — about $5\times$ faster run time than MCD.

### 4.3. Ablation Study

We also performed an ablation study to analyze several aspects of the proposed method. For computational efficiency, we ran all the studies over a randomly picked subset of the MCD dataset comprising 50 images. The results are reported below.

**Choices of the Blur Formation Model** $b(\hat{I}_S^{(t)}; K)$  In addition to the blur formation model described in Section 3.3, which is based on per-pixel convolution, one can warp the estimated sharp image $\hat{I}_S^{(t)}$ towards $t+1$ and $t-1$ directly using the the optical flow $\mathcal{F}_{t+1 \to t}$ and $\mathcal{F}_{t-1 \to t}$, and average the resulting images. The warping can be implemented with bilinear interpolation, making this formation model also differentiable.

Table 5 summarizes the results. As shown, the per-pixel convolution blur formation model performs slightly better in terms of PSNR and SSIM, while the warping-based blur formation runs much faster during the fine-tuning. Nevertheless, both blur formation models produce significant improvement over the two baseline methods.

**Self-supervised Learning on a Single Image** Since the proposed method is self-supervised, in theory one can fine-tune the deblur network for each individual image separately,

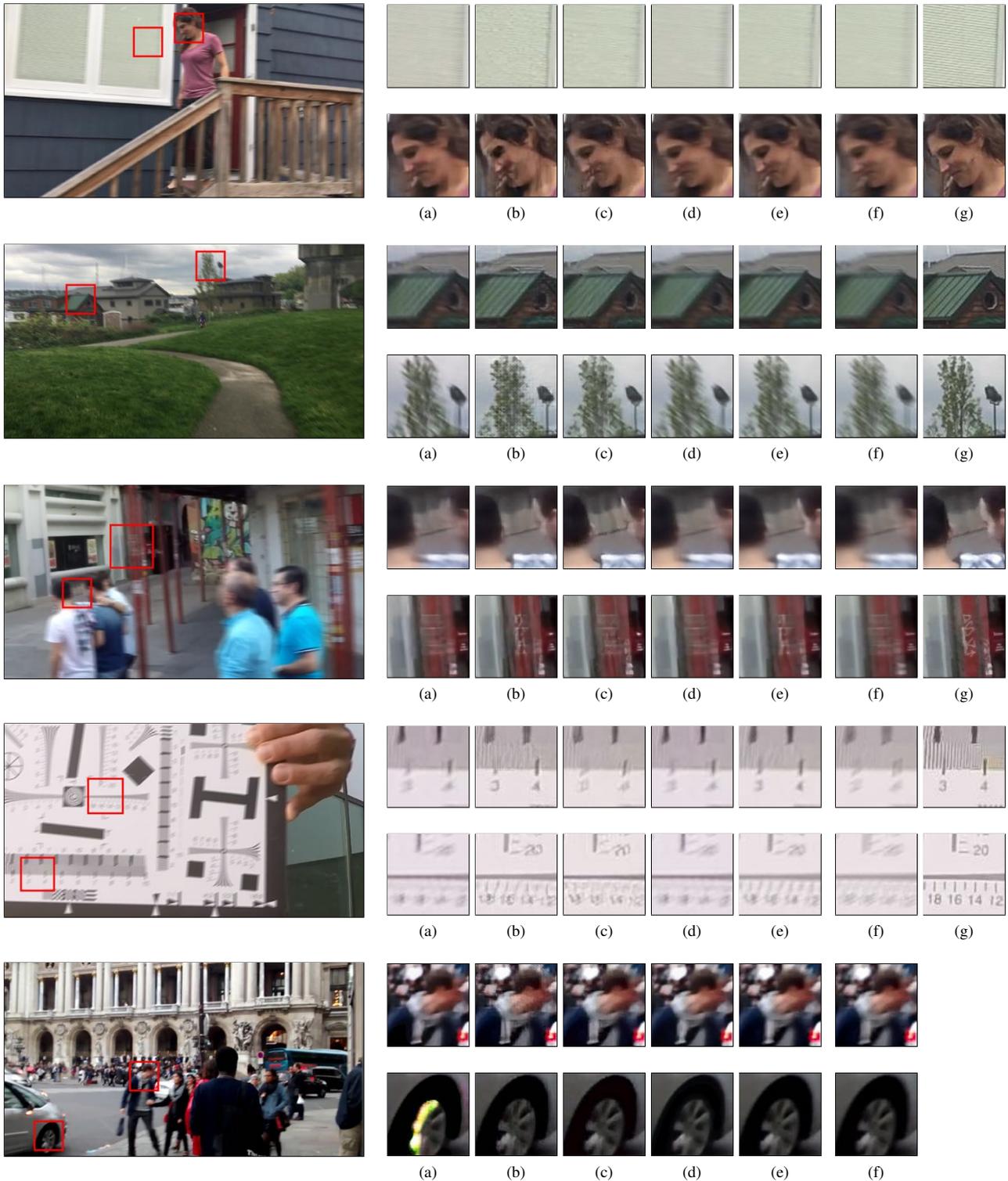

Figure 5. **Comparison of several deblur methods on images from different datasets**. The images are from DVD [22], MCD [16], WFA [4, 3] and our own ISOCHART datasets. The insets on the right shows the detailed input and deblur results of the bounding box areas in the left input images. The deblur results follow the order of (a) DVD baseline [22] (b) DeblurGAN [13], (c) MCD [16], (d) DVD-Reblur (ours), (e) DeblurGAN-Reblur (ours), (f) blurry (input), (g) ground truth. (The ground truth is not available for the WFA dataset, therefore the last column of the last row missing.) The complete, full resolution images are available in the supplementary material.

Table 2. Comparison of several state-of-art deblurring techniques (in terms of average PSNR, SSIM and average run-time) on three datasets. DVD-Reblur and DeblurGAN-Reblur are proposed methods which integrate the reblurring framework within existing DNN-based deblurring algorithms.

| Network→<br>Datasets↓ | DVD [22] PSNR | SSIM | DVD-Reblur PSNR | SSIM | DeblurGAN [13] PSNR | SSIM | DeblurGAN-Reblur PSNR | SSIM |
|---|---|---|---|---|---|---|---|---|
| MCD | 25.36 | 0.8380 | **26.06** | **0.8515** | 27.30 | 0.8954 | **28.03** | **0.9078** |
| DVD | 29.15 | 0.9218 | **30.15** | **0.9265** | 30.16 | 0.9364 | **31.37** | **0.9400** |
| ISOCHART | 29.85 | 0.9632 | **30.32** | **0.9706** | 31.89 | 0.9796 | **31.96** | **0.9814** |

Table 3. Results of two different blur formation models. The column "Time" is the average running time for fine-tuning a mini-batch of 10 RGB images of size $128 \times 128$.

| Network | PSNR | SSIM | Time (s) |
|---|---|---|---|
| DVD [22] | 25.067 | 0.872 | - |
| DVD-Reblur, Conv-Based | **25.694** | **0.880** | 2.15 |
| DVD-Reblur, Warp-Based | 25.693 | 0.880 | **0.56** |
| DeblurGAN [13] | 26.884 | 0.913 | - |
| DeblurGAN-Reblur, Conv-Based | **27.531** | **0.923** | 2.32 |
| DeblurGAN-Reblur, Warp-Based | 27.529 | 0.922 | **1.01** |

Table 4. Fine-tuning deblur network on single image vs. on group of images. Each approach's resulting PSNR and SSIM are shown.

| | Individual | | Group | |
|---|---|---|---|---|
| Image No. | PSNR | SSIM | PSNR | SSIM |
| 1 | 26.538 | 0.878 | **27.004** | **0.894** |
| 2 | 27.813 | 0.915 | **28.201** | **0.923** |
| 3 | **23.783** | 0.843 | 23.776 | **0.844** |
| 4 | 25.453 | 0.863 | **25.855** | **0.880** |
| 5 | 24.316 | 0.886 | **25.231** | **0.910** |

Table 5. Deblurring performance of three-frame based vs. two-frame based blur formation model.

| Network | PSNR | SSIM |
|---|---|---|
| DVD [22] | 25.067 | 0.872 |
| DVD-Reblur (2 frames) | 25.163 | 0.869 |
| DVD-Reblur (3 frames) | **25.694** | **0.880** |

## 5. Conclusion and Discussion

In this paper, we proposed a novel deep learning based method for video motion deblurring. In order to improve the generalization ability and overcome the image artifacts of prior supervised-learning based methods, we propose to incorporate a physics-based blur formation model to reblur the estimated sharp images, which allows us to fine-tune the deblur network via self-supervised learning. We evaluated our method over multiple datasets and found the proposed approach effectively removes image artifacts and improves performance.

There are several limitations in the current approach that we plan to address in the future. While the piece-wise linear blur kernel based on optical flow is applicable for most motion blur in videos, this assumption does not hold for large amount of motion blur which often results in nonlinear, complex blur kernels. Moreover, the GAN-based training has been shown to learn the statistical distribution of images well — it can be also incorporated with the physics-based blur formation model for the self-supervised learning.

despite the high computational cost. Interestingly, we found this customized, single-image based fine-tuning to be not necessarily better — slightly worse, in fact — than fine-tuning over a group of testing images. Table 4 summarizes the results for five randomly picked images, each of which is fine-tuned separately with DeblurGAN-Reblur. One possible explanation is that fine-tuning over a set of testing images may produce a more stable gradient based on the underlying image distribution, and may thus be less likely to get stuck in local minima. The full-resolution results of these five images are provided in the supplementary material.

**Effect of the number of frames in the blur formation model** When we construct the physics-based blur formation model, we need to compute two optical flow maps $\mathcal{F}_{t+1 \to t}$ and $\mathcal{F}_{t-1 \to t}$, which requires at least three frames as input. In addition, we also experimented with only two frames as input, and made the assumption that $\mathcal{F}_{t+1 \to t} \approx -\mathcal{F}_{t-1 \to t}$. We evaluated with DVD-Reblur network. As expected, we found that the three-frame based method performs better than the two-frame based method. Table 5 summarizes the results.